\documentclass{article}

\usepackage{microtype}
\usepackage{graphicx}
\usepackage{subfigure}
\usepackage{booktabs} 
\usepackage{tcolorbox}
\usepackage{lipsum,adjustbox}

\usepackage{hyperref}

\usepackage[accepted]{icml2023}

\usepackage{amsmath}
\usepackage{amssymb}
\usepackage{mathtools}
\usepackage{amsthm}

\usepackage[capitalize,noabbrev]{cleveref}

\theoremstyle{plain}

\theoremstyle{definition}

\theoremstyle{remark}

\usepackage[textsize=tiny]{todonotes}

\icmltitlerunning{LLM Guided Inductive Inference for Solving Compositional Problems}

\begin{document}
\usetikzlibrary{shapes.geometric, arrows}

\twocolumn[
\icmltitle{LLM Guided Inductive Inference for Solving Compositional Problems}



\icmlsetsymbol{equal}{*}

\begin{icmlauthorlist}
\icmlauthor{Abhigya Sodani}{yyy}
\icmlauthor{Lauren Moos}{zzz}
\icmlauthor{Matthew Mirman}{yyy}
\end{icmlauthorlist}

\icmlaffiliation{yyy}{Anarchy, Extensional, Inc, California, USA}
\icmlaffiliation{zzz}{Yoyodyne Inc, California, USA}
\icmlcorrespondingauthor{Abhigya Sodani}{abhigya@anarchy.ai}
\icmlcorrespondingauthor{Matthew Mirman}{matt@anarchy.ai}

\icmlkeywords{Machine Learning, ICML}

\vskip 0.3in
]



\printAffiliationsAndNotice{}  

\begin{abstract}
While large language models (LLMs) have demonstrated impressive performance
in question-answering tasks, their performance is limited when the questions require
knowledge that is not included in the model’s training data and can only be acquired
through direct observation or interaction with the real world. Existing methods decompose reasoning tasks through the use of modules invoked sequentially, limiting their ability to answer deep reasoning tasks. We introduce
a method, Recursion based extensible LLM (REBEL), which handles open-world, deep reasoning tasks by employing automated reasoning techniques like dynamic planning and forward-chaining strategies. 
REBEL allows LLMs to reason via recursive problem decomposition and utilization of external tools. 
The tools that REBEL uses are specified only by natural language description. We further demonstrate REBEL capabilities on a set of problems that require a deeply nested use of external tools in a compositional and conversational setting.
\end{abstract}

\section{Introduction}
\label{submission}

Recently, neural models for natural language generation have demonstrated impressive results \cite{koroteev2021bert, devlin2018bert, brown2020language}, opening significant new avenues for solving natural language reasoning tasks precisely \cite{huang2022towards, qiao2022reasoning}. While LLMs have shown a unique ability to scale in predictable and efficient ways, is unclear whether they show this scaling behavior on complex reasoning tasks \cite{huang2022towards}. Moreover, the limitations of large language models in accessing dynamic external knowledge sources significantly restrict their usefulness. Human reasoning involves a combination of observations and interactions with the world, highlighting the action-oriented nature of reasoning. 

In this paper, we address this by introducing the Recursion Based Extensible LLM (REBEL) framework. REBEL allows LLMs to reason through highly complex problems that require knowledge from disparate external sources. This is accomplished using an inference engine that, using the provided tools, gathers the necessary facts to infer the correct answer. Specifically we show three contributions:
\begin{enumerate}
    \item Designing a system capable of answering questions using any arbitrary external tool.
    \item An evaluation showing that REBEL improves upon the state-of-the-art performance on multi-Hop fact retrieval and compositional question answering problems.
    \item Releasing our code and evaluation suite for open-source usage at \url{rebel.anarchy.ai}.
\end{enumerate}

\section{Related Works}

At a high-level, methods for approaching reasoning tasks using LLMs can be broken down into prompt engineering techniques \cite{liu2023pre, llmprogramming} and fine-tuning \cite{micheli2021language, schick2023toolformer} techniques, or combinations of the above. Here we focus only on prompt techniques. 

Forward chaining \cite{liebowitz1988introduction} is a reasoning strategy historically used by expert systems. It operates by repeatedly applying logical inference rules from an initial repository of known axioms to eventually ideally produce the goal. This strategy has recently been employed to solve natural language problems with the assistance of LLMs in Chain of Thought (CoT) \cite{CoT}.  ReAct \cite{react} builds off of CoT by generating task-specific actions in response to reasoning. Chameleon \cite{lu2023chameleon} takes this further, using LLMs to synthesize tool pipelines including off-the-shelf computer vision models, web-search engines, and calls to generative models.  In contrast to forward-chaining, the technique of backward-chaining \cite{russell2010artificial} attempts to limit the search-space of possible inferences by determining what must be true for a goal to be shown \cite{picco2021neural}.  

\citet{selfask} demonstrates a method to evaluate problem-solving abilities on a category of non-trivial reasoning tasks with {\em compositional} structure \cite{lake2018generalization, keysers2019measuring} that is poorly addressed by prior methods. They express compositional error as the number of questions in which two subquestions are answered correctly but the top-level question is not.  
Prior work has shown how this can be addressed via problem decomposition \cite{yang2022seqzero, zhou2022least, drozdov2022compositional, khot2022decomposed}.  
In this work, we show how problem decomposition can be augmented with tool usage.

\section{Methods}
\begin {figure*}
\label{Figure 1}
  \caption{Visual depiction of the pipeline of the REBEL Algorithm from \cref{alg:1} to answer some $Question_n$. Blue boxes contain descriptions of each step of the pipeline,  and the red boxes contains the output variable for each step that will be used in subsequent steps.
  \vspace{0.2cm}}
\centering
\begin{adjustbox}{width=\textwidth}
\begin {tikzpicture}[node distance=1.5cm]
\linespread{0.2}

\tikzstyle{startstop} = [rectangle, rounded corners=1px,  
minimum height= 0.1cm,
font = \tiny,
text centered,
text width = 1.05cm,
draw=none, 
fill= none]

\tikzstyle{stage} = [rectangle,  rounded corners=1px,
minimum height= 0.8cm,
font = \tiny,
text centered,
text width = 1.05cm,
draw=black, 
fill=blue!30]

\tikzstyle{output} = [rectangle, rounded corners=1px, 
minimum height= 0.5cm,
font = \tiny,
text centered,
text width = 0.9cm,
draw=black, 
fill=red!30]

\tikzstyle{arrow} = [->,>=stealth]

\node (0) [startstop] {\fontsize{3pt}{1}\selectfont Question Splitting};
\node (label0) [stage, below of=0, yshift=1.0cm] {\fontsize{2.4pt}{1}\selectfont 1) Generates \textbf{Subquestions\textsubscript{n}} and recursively gets their answers and appends them to \textbf{Memory\textsubscript{n}}.};
\node (output0) [output, below of=label0, yshift=0.90cm] {\fontsize{3pt}{2}\selectfont \textbf{Memory\textsubscript{n}}};

\node (1) [startstop, right of=0] {\fontsize{3pt}{1}\selectfont Memory Checking};
\node (label1) [stage, below of=1, yshift=1.0cm] {\fontsize{2.4pt}{1}\selectfont 2) Generates a boolean \textbf{MemoryCheck\textsubscript{n}} which indicates if the answer to \textbf{Questions\textsubscript{n}} can be found in \textbf{Memory\textsubscript{n}}.};
\node (output1) [output, below of=label1, yshift=0.90cm] {\fontsize{3pt}{2}\selectfont \textbf{MemoryCheck\textsubscript{n}}};

\node (2) [startstop, right of=1] {\fontsize{3pt}{1}\selectfont Tool Picking};
\node (label2) [stage, below of=2, yshift=1.0cm] {\fontsize{2.4pt}{1} \selectfont 3) If \textbf{MemoryCheck\textsubscript{n}}is false, we generate \textbf{ Tool\textsubscript{n}}, the best tool to answer \textbf{Question\textsubscript{n}}};
\node (output2) [output, below of=label2, yshift=0.90cm] {\fontsize{3pt}{2}\selectfont \textbf{Tool\textsubscript{n}}};

\node (3) [startstop, right of=2] {\fontsize{3pt}{0.5}\selectfont Tool Input Generation};
\node (label3) [stage, below of=3, yshift=1.0cm] {\fontsize{2.4pt}{1} \selectfont 4) Takes \textbf{Memory\textsubscript{n}} and \textbf{Question\textsubscript{n}} and generates input for \textbf{Tool\textsubscript{n}}};
\node (output3) [output, below of=label3, yshift=0.90cm] {\fontsize{3pt}{2}\selectfont \textbf{... "tool\textsubscript{n}param\textsubscript{k}": "tool\textsubscript{n}value\textsubscript{k}", … }
};
\draw [arrow] (label0) -- (label1);
\draw [arrow] (label1) -- (label2);
\draw [arrow] (label2) -- (label3);
\end{tikzpicture}
\end{adjustbox}
\end{figure*}
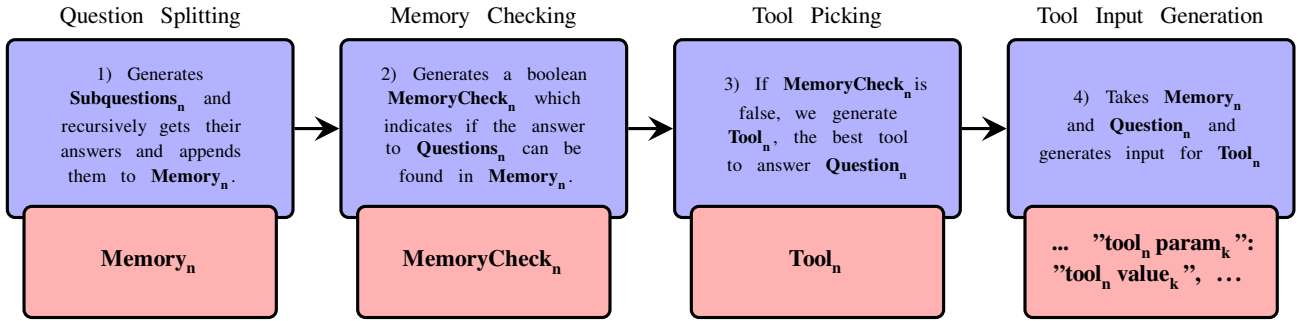

In this section we introduce the REBEL algorithm as shown in Fig.~\ref{alg:1}, and all necessary notation and background. At a high level, it works recursively to solve questions, breaking questions into subquestions until no further subquestions can be generated. Let us call the $nth$ question/subquestion $Question_n$ and its answer $Answer_n$. For example, the user-provided question would be $Question_0$. Let us call the subquestions that are generated to answer $Question_n$ $Subquestions_n$. In each recursive step, we break $Question_n$ into $Subquestions_n$. Let us call the answer to the $ith$ member of $Subquestions_n$ $subanswers_n[i]$. We recursively call each member of $Subquestions_n$, and each $subanswers_n[i]$ is returned as a $fact$ which is the tuple $(Subquestions_n[i],subanswers_n[i])$. This fact is appended to a list of $facts$ that is is global to each $Question_n$.  This list of $facts$ becomes $Memory_n$ which is used to inform $Answer_n$.

In order to stop unbounded recursion, we delete members of $Subquestions_n$ whose featurizations have cosine similarities above 0.98 to the featurization of $Question_n$.

The REBEL system contains a $Tool \textunderscore List$, which is a numbered list of the tools we have available and their descriptions. If required, we determine a $Tool_n$ for each $Question_n$, which is the number of the tool  required to answer $Question_n$ given $Memory_n$.

Below we define the basic steps of this algorithm:  question splitting, checking memory, picking tools, and using tools. Figure 1 depicts this pipeline. 
\begin{algorithm}[tb]
   \caption{REBEL}
   \label{alg:1}
\begin{algorithmic}
\FUNCTION{promptf($Question_n$,$facts$,$allowsplit=True$)}
   \IF{$\text{$allowsplit$}$}
        \STATE $Subquestions_n$ = split($Question_n$,   $facts$) \COMMENT{split the question into subquestions to answer}
        \FOR{$subquestion$ from 1 to $s$ in $Subquestions_n$}
            
            \IF{$\text{cos\textunderscore similarity($Question_n$,$subquestion$) $>$ 0.98}$}
                \STATE Delete $subquestion$
                \STATE $allowsplit$=False
            \ENDIF
        \ENDFOR
        \FOR{$subquestion$ from 1 to $s$ in $Subquestions_n$}
        \STATE \textunderscore,$newfacts$=\textsc{promptf}
        \STATE(\text{$subquestion,facts,allowsplit$})
        \STATE $facts$ += $newfact$
        \ENDFOR
    \ENDIF
    \IF{$\textsc{memorycheck}(Question_n,facts)$}
        \STATE $Answer_n=\textsc{callGPT}(Question_n,facts)$
        \STATE \textbf{return} $Answer_n,(Question_n,Answer_n)$
    \ELSE
        \STATE $tool = \textsc{pick tool}(Question_n,facts)$
        \STATE $toolinput = \textsc{callGPT} (tool,Question_n,facts)$ \COMMENT {to determine tool input}
        \STATE $Answer_n = \textsc{useTool}(toolinput,facts)$
    \ENDIF
    \STATE \textbf{return} $Answer_n$,$(Question_n,Answer_n)$
   \ENDFUNCTION
\end{algorithmic}
\end{algorithm}
\subsection{Question Splitting}

The split subroutine divides $Question_n$ into $Subquestions_n$ with the size of $Subquestions_n$ being the number of subquestions that the LLM generates. The LLM is prompted with $Tool\textunderscore List$, and 4 shots of question splitting examples. This step is representing in step 1 of Figure 1. To see a single shot of context for question splitting see \cref{Appendix A}.

We answer each subquestion and its results are returned as a $fact$ (see \cref{alg:1}). These facts are accumulated and passed to all subsequent subquestions. The list $Subquestions_n$ is ordered such that the $fact$ gained from answering a lower indexed subquestion will aid in the answering of a higher indexed subquestion. 
\subsection{Memory Check}

We check if a question can be answered without any tool use. This can mean either that the question can be answered using $Memory_n$ or the question can be answered by an LLM without the use of any tools (see step 2 Figure 1). If this is the case, we directly provide our base LLM $Memory_n$ and $Question_n$ to find $Answer_n$. To see the complete memory check prompt see \cref{Appendix B}.

\subsection{Tool Picker}
Here we evoke the LLM to decide what member of $Tool\textunderscore List$ (described by integer $Tool_n$) would be best to decide the answer to a question. This is a 0-shot prompted system which can be seen in step 3 of Figure 1.

\subsection{Tool Input Generation}
We use GPT-3 to generate standardized input to our tools. We provide the tools to the LLM with 2 fields. The description of the tool and the dynamic parameters of the tool. We store the 3 more fields about each tool that are hidden from the LLM, these are: if the tool is a GET/POST request, the endpoint URL for the tool, and the static parameters of the tool. The dynamic parameters are parameters that will be adjusted based on each call (for example, and query field). The static parameters are parameters that stay the same on each API call (for example, an authentication key). 

REBEL uses 3 default tools: search, weather, and Google Maps. We configure the inputs to every tool as a JSON. A tool input JSON maps a given tool's dynamic parameters to the values those parameters should be in order to obtain information to answer a given question: $\{"tool_nparam_1": "tool_nvalue_1",$ ...
$"tool_nparam_k":"tool_nvalue_k"\}$.
 A standardized JSON format reduces the load on the LLM to format an entire API call by itself.  

REBEL allows for an arbitrary tools to be added to it, however, the k-shot examples that are provided to the LLM for generating input given $Tool_n$ are designed around the 3 base tools. We have found that this prompting does extrapolate to 0-shot uses of unseen and arbitrary tools. See \cref{Appendix C} for a complete single shot of tool input generation context.

\subsection{Tool Usage}
The UseTool function takes the dynamic parameters (from the LLM generated tool input), the static parameters that we have stored for each tool, and the API endpoint and makes a single request URL. This URL is requested, and the return output is stored as a string. If the return output is longer than 15,000 characters, it is truncated to that amount. Then, we use an LLM, provided with $Memory_n$, $Question_n$, and the API request output, to generate an answer to the $Question_n$. This answer is returned from the UseTool function as $Answer_n$. Our approach has some consequences. On the positive side, users do not have the indicate how to parse the output of the tools they give us, this makes REBEL extremely extendable and flexible to interpret many tool return types and formats. On the negative side, because of the extremely unstructured nature of tool returns, errors are caused by UseTool not being able to answer a question based on a tool return.

\section{Evaluation}

In this section we first introduce the experimental setup, including the benchmarks used for evaluation, and then present the results.  

\subsection{Experimental Setup}

We tested REBEL on 3 datasets: Compositional Celebrities \cite{selfask}, FEVER \cite{thorne2018fever}, and HotPotQA \cite{yang2018hotpotqa}.

On these datasets, correctness was determined by a human experimenter based on the output of each system. ReAct outputs with simply the answer to the question, while REBEL often outputs the answer wrapped in reasoning behind the system's thoughts. For these experiments, two separate sets of rules had to be determined for fact verification and fact retrieving questions. For fact retrieving questions, an answer was considered correct if the desired answer was contained in the system output. For fact verification, if the model output determination of the truthfulness of a statement was the same as the desired truthfulness, then the generated answer was considered correct. 

On Compositional Celebrities, due to computational limitations, we tested using 5 of the 17 categories available,  using 100 questions per category, randomly chosen. These categories can be found in \cref{table1}.

We tested on FEVER and HotPotQA with 100 of the same random questions from each dataset on both ReAct and REBEL. The accuracy results for this experiment can be found at \cref{table2}. FEVER has 3 types of potential output labels (SUPPORTS, REFUTES, NOT ENOUGH INFO). In order to make prevent accidental correct answers from the REBEL system, only questions with the SUPPORTS and REFUTES labels were considered. 

For this experiment REBEL was only allowed to use a search tool to query the internet, as that is the only tool that the ReAct system has access to. 

Our code, which can be found at \url{rebel.anarchy.ai}, was implemented in Python using the OpenAI Completion API to access GPT-3 (text-davinci-003).

\subsection{Results}
We found that REBEL outperformed ReAct on answering questions that require i) the gathering of many facts to determine an answer ii) very specific search queries that return large amounts of unstructured data. With our experimental results we were able to show that REBEL is a state-of-the-art system in terms of its ability to consistently answer questions from disparate knowledge bases.

\subsubsection{Multi-Hop Fact Retrieval}
We used 2 datasets to test multi-hop fact retrieval: Compositional Celebrities and HotPotQA. 

Compositional Celebrities is a dataset consisting of 8.6k questions about Celebrities in different categories. All questions require retrieving two facts and basic reasoning. These two facts have never co-occurred in any text that would conceivably be part of the LLM training and the only way that the conclusion could be reached is for both of them to be evaluated correctly and composed with one another. We found that the REBEL system largely outperformed the ReAct system at all of the 5 categories that were experimented on for Compositional Celebrities. \textbf{On average, over the 5 categories tested, REBEL beat ReAct by 27.6 percent.} The reason for this is likely the ability of the REBEL system to work with unstructured tool return data. This allows the REBEL system to make and interpret very specific tool queries, whereas other systems that require standardized output can become constricted by the by a smaller possible set of tool queries. The results of this experiment can be found in \cref{table1}. 
\\
\\
HotpotQA is a challenging question-answering dataset containing 113,000 pairs of questions and answers derived from Wikipedia articles. The questions in HotpotQA necessitate synthesis of information from diverse sources and cannot be found pre-existing training knowledge bases. \textbf{ReAct outperformed REBEL on HotPotQA by 13 percent} (\cref{table2}). 

HotPotQA has questions that are significantly more than 2-hops, and on these questions REBEL tends to generate a massive recursive tree of subquestions. This introduces the issue of generating subquestions that lose context of the original question. Many times this can lead to the LLM not being able to reason through the large context window generated when processing these layers of recursive subquestions, resulting in the LLM finding no solution.

\subsubsection{Fact Verification} To test fact verification abilities, we employed the FEVER dataset. This benchmark is designed to evaluate the ability of models to extract factual information from textual sources and verify claims. The fact verification task involves determining the accuracy of claims made in a given piece of text.

\textbf{On FEVER, the REBEL system (78 percent accuracy) performed slightly better (\cref{table2}) than ReAct system (72 percent).} The reason for this out-performance by the REBEL system is because of the significant amount of "facts" that it gathers during its recursive solving of a fact verification problem. On several occasions, the ReAct system cannot find the information it is looking for to answer a questions, and therefore reports that it cannot make a determination if a certain fact is true or not.  

\begin{table}
\caption{Accuracy (percent of questions answered correctly) of different algorithms on the categories of Compositional Celebrities.}
\label{table1}
\vskip 0.15 in
\begin{center}
\begin{small}
\begin{sc}
\begin{tabular}{lcccr}
\toprule
Category & ReAct & REBEL \\
\midrule
    birthplace \textunderscore rounded \textunderscore lat & 28 & \textbf{59}\\
    birthplace \textunderscore currency & 85 & \textbf{94} \\
    birthplace \textunderscore currency \textunderscore symbol & 35 & \textbf{47} \\
    birthyear \textunderscore nobelLiterature & 33 & \textbf{82} \\
    birthdate \textunderscore uspresident & 53 & \textbf{90} \\
   \bottomrule
\end{tabular}
\end{sc}
\end{small}
\end{center}
\vskip -0.1in
\end{table}

\begin{table}
\caption{Accuracy (percent of questions answered correctly) of different algorithms on HotPotQA and FEVER.}
\label{table2}
\vskip 0.15in
\begin{center}
\begin{small}
\begin{sc}
\begin{tabular}{lcccr}
\toprule
Dataset & ReAct& REBEL \\
\midrule
    FEVER & 72 & \textbf{78}\\
    HotPotQA & \textbf{63} & 50 \\
   \bottomrule
\end{tabular}
\end{sc}
\end{small}
\end{center}
\vskip -0.1in
\end{table}

\begin{table*}[t]
\caption{Accuracy (percent of questions answered correctly) of different algorithms on the categories of Compositional Celebrities.}
\label{table3}
\vskip 0.15in
\begin{center}
\begin{small}
\begin{sc}
\begin{tabular}{lcccr}
\toprule
Category & GPT3 & REBEL w/o tools & REBEL  \\
\midrule
    birthplace \textunderscore rounded \textunderscore lat & 16 & 39 & \textbf{59}\\
    birthplace \textunderscore currency & \textbf{95} & 94 & 94 \\
    birthplace \textunderscore currency \textunderscore symbol & 28 & 45 & \textbf{47} \\
    birthyear \textunderscore nobelLiterature & \textbf{95} & 90 & 82 \\
    birthdate \textunderscore uspresident & 44 & \textbf{91} & 90 \\
   \bottomrule
\end{tabular}
\end{sc}
\end{small}
\end{center}
\vskip -0.1in
\end{table*}

\begin{table*}[t]
\caption{Accuracy (percent of questions answered correctly) of different algorithms on HotPotQA and FEVER.}
\label{table4}
\vskip 0.15in
\begin{center}
\begin{small}
\begin{sc}
\begin{tabular}{lcccr}
\toprule
Dataset & GPT3 & REBEL w/o tools & REBEL \\
\midrule
    FEVER & 77 & 73 & \textbf{78}\\
    HotPotQA & 43 & 46 & \textbf{50} \\
   \bottomrule
\end{tabular}
\end{sc}
\end{small}
\end{center}
\vskip -0.1in
\end{table*}
\subsection{Ablation Study}
In order to determine the efficiency of REBEL, we conducted several ablation tests. In these tests the aim was to isolate the affect of the REBEL system upon compositional problem solving. We used plain GPT3 (text-davinci-003) as our baseline. The results of these tests are in (\cref{table3} and \cref{table4}).

These tables show that GPT3 outperforms REBEL (with or without an external search tool) when a question can be easily answered with data that GPT3's training set. This is seen in \cref{table3} in the rows pertaining to $Birthyear\textunderscore NobelLiterature$ and $Birthplace \textunderscore Currency$. 

The REBEL algorithm without the external search tool outperformed the baseline when information processing is necessary to determine a final answer. Examples of this include questions that required the returning of a currency symbol or a rounded latitude. GPT3 succeeded in fetching the currency name or latitude correctly, but failed to round the latitude or return the symbol associated with the currency name. Adding external search augmented the REBEL algorithm's ability to reason with current facts, and therefore furthered the REBEL algorithm performance on most categories of Compositional Celebrities. Occasionally, the inclusion of an external search tool decreased performance due to the unstructured nature of return data the external tool provided. An example of this is on the $Birthyear\textunderscore NobelLiterature$ category of Compositional Celebrities. 

On most categories of Compositional Celebrities and on HotPotQA, REBEL without the use of an external search tool improved performance over baseline GPT3. This indicates that our recursive approach adds reasoning capability to GPT3 independently of external tool use. 

\begin{table}[t]
\caption{Average time taken to answer a question from Compositional Celebrities}
\label{table5}
\vskip 0.15in
\begin{center}
\begin{small}
\begin{sc}
\begin{tabular}{lcccr}
\toprule
Algorithm & Time (s) \\
\midrule
    GPT3 & 0.94 \\
    REBEL w/o tools & 5.358  \\
    REBEL w/ tools & 9.76 \\
   \bottomrule
\end{tabular}
\end{sc}
\end{small}
\end{center}
\vskip -0.1in
\end{table}
\section{Cost Analysis}
The recursive search nature of the REBEL algorithm means that it employs many calls to an LLM before determining an answer to a question. The downsides of this approach manifest themselves in latency (\cref{table5}) and monetary cost of LLM queries. Any external tools that are provided to the REBEL system will also be called very frequently, potentially leading to REBEL being a monetarily expensive system on that front as well. 
\\
\\
If a user desires to use REBEL without any tools, a cost in terms of hallucination has a potential of arising. Due to the lack of any external knowledge base, a hallucination on one subquestion has the potential to pollute the entire tree of reasoning. 
 
\section{Conclusion}
We have introduced REBEL, a recursive reasoning algorithm designed to use any arbitrary API as an external tool. REBEL outperforms the state-of-the-art on questions that require the collection of many facts and those that benefit from the ability to make highly specific queries to outside sources of data, which may be unstructured. REBEL also has a demonstrable improvement over the GPT3 LLM when answering questions that require multi-step information processing. However, the REBEL algorithm tends to over-complicate simple problems, leading to a reduction in accuracy when compared to baseline GPT3 on questions that require minimal compositionality.

Future work would ideally address fine-tuning LLMs for each step in the REBEL pipeline and experimenting with limiting recursive depth of subquestion generation.  
\bibliography{paper}
\bibliographystyle{icml2023}

\newpage
\onecolumn
\appendix
\section{Appendix A}
\textbf{Question Splitting Prompt}
\label{Appendix A}
\begin{tcolorbox}

Tools we have access to =

tool 1: The tool returns the results of free-form queries similar to those used for wolfram alpha. This is useful for complicated math or live data retrieval.  Can be used to get the current date.

tool 2: Find the driving distance and time to travel between two cities.

tool 3: Find the weather at a location and returns it in celcius.

Q=\textbf{$Question_n$}

Look at the tools we have access to. Split Q into subquestions to answer Q that can each be solved with one use of one tool. Make as few subquestions as possible. Split each subquestion with a comma and have no extra information other than the subquestions.

\end{tcolorbox}

\section{Appendix B}
\textbf{Memory Check Prompt}
\label{Appendix B}
\begin{tcolorbox}

Q: "What's the time?"
Is the answer to Q found in the memory or in your knowledge base already? Answer with a yes or no. no

Q: "How you feeling?"
Is the answer to Q found in the memory or in your knowledge base already? Answer with a yes or no. yes

Q: "What color is the sky"
Is the answer to Q found in the memory or in your knowledge base already? Answer with a yes or no. yes

Q: "What is the temperature in Portland?"
Is the answer to Q found in the memory or in your knowledge base already? Answer with a yes or no. no

Memory: \textbf{$Memory_n$}

Q: \textbf{$Question_n$}
Is the answer to Q found in the memory or in your knowledge base already? Answer with a yes or no.

\end{tcolorbox}

\section{Appendix C}
\textbf{Tool Input Prompt}
\label{Appendix C}
\begin{tcolorbox}
\begin{verbatim}
<TOOL>
<ID>1</ID>
<DESC>Find the driving 
distance and time to 
travel between two 
cities.</DESC>
<PARAMS>{"origins": the origin city,
"destinations": the destination
city}</PARAMS>
</TOOL>

<CASE>
<Q>How long would it take 
to get between 
South Africa and Kenya.
</Q>
<THOUGHT>
<P>What 
should the input for 
tool 1 be to 
answer Q?</P>
<A ty=JSON>
{"origins": "South Africa", 
"destinations": "Kenya"}
</A>
</THOUGHT>
</CASE>
\end{verbatim}
\end{tcolorbox}

\end{document}